%% file: main.tex
\documentclass[letterpaper, 10 pt, conference]{ieeeconf}  
\renewcommand{\baselinestretch}{0.98}

\IEEEoverridecommandlockouts                              

\usepackage{todonotes} 
\usepackage{graphicx}  
\usepackage{xcolor}

\usepackage{amsmath}  
\usepackage{amssymb}  

\usepackage{hyperref}
\usepackage{scalefnt}

\usepackage[final]{changes}

\newcommand{\phm}{\phantom{-}}

\input{mathdef.tex} 

\title{\LARGE \bf Predicting the Post-Impact Velocity of a Robotic Arm \\ via Rigid Multibody Models: an Experimental Study}

\author{Ilias Aouaj$^{1}$, Vincent Padois$^{2}$, Alessandro Saccon$^{3*}$
\thanks{This work was partially supported by the  Research  Project  I.AM. through  the  European  Union H2020 program under GA 871899 .}
\thanks{$^{1}$Ilias Aouaj has conducted this work during his MSc project at the Department of Mechanical Engineering, Eindhoven University of Technology (TU/e), The Netherlands
({\tt\small aouajilias@gmail.com})
}
\thanks{$^{2}$Vincent Padois is with Auctus, Inria - IMS (Univ. Bordeaux / Bordeaux INP / CNRS UMR 5218), 33405 Talence, France
({\tt\small vincent.padois@inria.fr})
}
\thanks{$^{3*}$Alessandro Saccon (corresponding author) is with the Department of Mechanical Engineering, Eindhoven University of Technology (TU/e), The Netherlands
({\tt\small a.saccon@tue.nl})
}%
}

\begin{document}
\scalefont{0.999}

\maketitle
\thispagestyle{empty}
\pagestyle{empty}


\begin{abstract}
Accurate post-impact velocity predictions are essential in developing impact-aware manipulation strategies for robots, where contacts are intentionally established at non-zero speed mimicking human manipulation abilities in dynamic grasping and pushing of objects.
Starting from the recorded dynamic response of a 7DOF torque-controlled robot that intentionally impacts a rigid surface, we investigate the possibility and accuracy of predicting the post-impact robot velocity from the pre-impact velocity and impact configuration. The velocity prediction is obtained by means of an impact map, derived using the framework of nonsmooth mechanics, that makes use of the known rigid-body robot model and the assumption of a frictionless inelastic impact.

The main contribution is proposing a methodology that allows for a meaningful \emph{quantitative} comparison between the recorded post-impact data, that exhibits a damped oscillatory response after the impact, and the post-impact velocity prediction derived via the readily available rigid-body robot model, that presents no oscillations and that is the one typically obtained
via mainstream robot simulator software.
The results of this new approach are promising in terms of prediction accuracy and thus relevant for the growing field of impact-aware robot control. The recorded impact data (18 experiments) is made publicly available, together with the numerical routines employed to generate the quantitative comparison, to further stimulate interest/research in this field.
\end{abstract}

\section{INTRODUCTION}
State-of-the-art robot manipulation is  performed by intentionally establishing contact at almost-zero velocity~\cite{SaBi18J_DS_contact_transition}. While this strategy is effective
and provides guarantees regarding the successful accomplishment of the task, it also increases  execution time as well as energy expenditure due to the associated acceleration/deceleration phases required to bring the contact velocity to zero when compared with a scenario where contact is established at non-zero velocity. Modern robots, in particular torque-controlled robots,
are more and more designed for physical interaction,
by means of employing back drivable and compliant joints which provide
an impact torque ﬁltering functionality
and a
physical protection to the reduction drives
\cite{ASBi20B_SoftRobotics,TsetalJ17_Walkman, EnWeOtHeRoGaBuBeEiScAS14J_TORO}.
This allows for the experimentation of dynamic contact tasks where contact is established at non-zero velocity, leading to collisions which cause rapid changes in the system velocities and short-lived post-impact vibrations.

\begin{figure}[!ht]
    \centering
     \includegraphics[scale = 0.8]{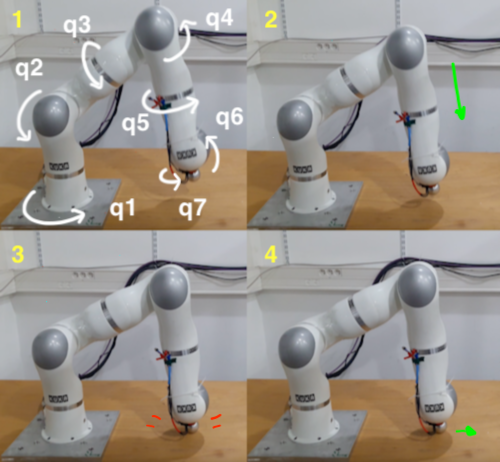}
    \caption{Example of a vertical impact test at 0.2 m/s conducted
    on a KUKA LWR IV+ robotic arm.
    The figure shows four snapshots from a recorded video.
    Time evolves from left to right and
    from top to bottom.
    \added{Starting from a rest pose (1),
    also illustrating the robot 7 DoFs, the robot
    moves down (2). After impact (3), the end-effector
    slides on the table (4), keeping contact.}.
    }
    \label{fig:snap_kuka_exp}
\end{figure}

To achieve complex dynamic contact tasks, in parallel to adequate compliant mechanical design,
it is necessary to develop a coherent
framework that encompasses control,
learning, planning, and sensing strategies
in the presence of impacts and sudden velocity jumps
\cite{RiMoTrNoWoSaNi17C_humanoidImpact, WaKh19C_ImpactFriendlyRobustCtrl}. These methods assume to have at hand reliable forecast
of the effects of impact phenomena
to be able to cope with expected (and unexpected) impacts and ensure stability and desired level of performance while executing impact tasks.

Given the availability of new generation robots
and everlasting desire/need to increase motion capabilities of robotic systems, the development
of impact models is steadily gaining importance in the robotics community \cite{HaPo19J_NonUniqueMultipleImpacts},
together with the need for validation of these models against real experiments  to assess their prediction ability and range of applicability \cite{FaZaDrRo17C_Limitations}. One can speculate that collision exploitation will become as important in robotics as traditional collision avoidance.

Robot-environment collision models have a long history in robotics \cite{ZhHe85J_MathModelingCollision, MI90C_ManipulationTransition,
HuMa94J_PlanrKinematicChainMultipleContact,
Wa94J_ImpactMeasures,
NeYo99J_ImpactAnalysisSpaceRobot}\added{, with important
contributions still appearing nowadays \cite{JiGaMu19J_Batting}}.
Despite this long history,
experimental validation of these collision models
is currently limited to robot locomotion \cite{PaSrHuGr11J_IDBipedalRobot},
single free-falling objects (typically, lightweight small spherical objects, cylinders, or dumbbells)
impacting a rigid surface or a robot manipulator, but
with limited/no effect on the
robot dynamics\deleted{\cite{JiGaMu19J_Batting}}. We are also unaware of publicly available robot-object-environment impact motion databases.

Acknowledging the emergence of robots capable of dynamic physical interactions,
provided with high-frequency
joint torque sensors and high-resolution encoders,
the long term vision of our investigation is instead that of {\em impact-aware manipulation}: we envision robots grabbing and pushing massive objects that have a fast and direct influence on the robot dynamics right after contact is established.
In this context, it is fundamental to be able to predict, both from a planning and control point of view, post-impact velocities \cite{RiSaNi20J_RS_1DOF}. This,
together with the creation of a publicly available repository of robot-object-environment impact motions, is also the goal of a recently started H2020 EU project\footnote{Impact-Aware Manipulation by Dexterous Robot Control and Learning in Dynamic Semi-Structured Logistic Environments. Project Website: \url{https://i-am-project.eu}}.

The aim of this paper is to fill in some of the gaps of the robotic literature identified above, first by making available\footnote{The impact data and associated MATLAB scripts (that make use of the Robotics Toolbox \cite{Co17B_RoboticsToolbox} for handling of the robot kinematics), can be downloaded from \cite{Dataset21D}.} the joint data of the torque-controlled robotic arm while
it impacts a wooden table
at different speeds and impact angles (see Figure~\ref{fig:snap_kuka_exp} for an illustration of this impact scenario \added{and Figure~\ref{fig:impact_table_0p2_00} for corresponding joint displacements}) and then proposing a \added{thinking} framework \added {and associated procedure} for quantitatively assessing the prediction ability of an impact map, derived from a known rigid body robot model, in determining the post-impact robot velocity.

As known from the field of nonsmooth mechanics, the impact map is the result of the combination of an impact equation and an impact law. For a robot manipulator, the impact equation is derived from an identified rigid body model or readily derived from 3D CAD design. When performing impact experiments, however, clear oscillatory transients are visible: in our experiments, these last for about a hundred milliseconds. At first, it is unclear how to assess the post-impact prediction performance of an impact map derived from
an idealized algebraic impact law
with
an idealized rigid-body robot model (the one available with state-of-the-art robot identification techniques \cite{HoKhGa16B_robotID, DLBo16B_FlexRobots}) versus the measured impact response. Hence, the main contribution of this work is to propose a methodology to separate the rigid response from the oscillatory response of the robot and consequently a methodology to assess the quality of the post-impact prediction.
\added{To the best of the authors knowledge,
this idea is new and cannot be
directly related to previously published work.}
Such a post-impact velocity prediction has value for planning robot-object motions with impacts as well as for developing collision monitoring strategies to discriminate between planned and unplanned collisions. The procedure is illustrated first on an academic example in one dimension and then applied to the recorded impact data of a 7DOF robotic arm, showing good prediction capability.

The paper is organized as follows.
Besides this introduction,
Section~\ref{sec:nonsmooth} provides
 basic background information regarding
nonsmooth mechanics and impact maps.
A description of the impact experiments and problem
statement is provided in Section~\ref{sec:probstatement}.
Section~\ref{sec:contribution} details the main contribution, which is first introduced for an easily accessible academic example. Application of the proposed methodology
on the 7DOF robot data is presented
in Section~\ref{sec:results}. Finally, conclusions
and future work is discussed in Section~\ref{sec:conclusions}.


\section{NONSMOOTH MECHANICS IMPACT MAPS}
\label{sec:nonsmooth}

Within dynamical systems theory, nonsmooth mechanics \cite{Br16B_nonsmooth} is
quite a mature theoretical framework that combines rigid body modeling with algebraic impact laws, with the aim of capturing the post-impact state of
a mechanical system based on the ante-impact configuration and velocity.
The essential modeling assumption within this framework
is a space-and-time scale separation
between the contact and body dynamics that justifies approximating the impact dynamics as instantaneous (i.e., taking zero time) and consequently allowing for instantaneous jumps in the system's velocity and corresponding impulsive contact forces.

While admitting instantaneous velocity jumps and impulsive contact forces is a clear idealization of the contact dynamics (for the family of robots we are considering in this work,
impact duration is typically in the range of 5 to 10 ms
as shown in \cite{HaASHi07C_SafetyEvaluation, HaASHi09J_RequirementsForSafeRobots, HaASDLHi_08C_CollisionDetectAndReact}), advanced impact models can provide impressive prediction capabilities even in the presence of multiple simultaneous impacts \cite{NgBr18J_simultineousImpact}.
Also, these algebraic impact models have demonstrated extremely effective in planning and control for mechanical systems undergoing impacts, going from the estimation of the distribution of possible poses of a known object dropped on a surface from an arbitrary height \cite{MoEr02J_PoseDistributions}, model-based dynamic robot locomotion \cite{GrChSiAm14J_3DBipedWalk, ReCoHeHuAm16C_DURUS}, and accurate batting of flying objects  \cite{JiGaMu19J_Batting}. There is therefore good hope
that similar models, once validated, can be
also of great use
in impact-aware robot manipulation.

In the simplest form assumed in this paper, the {\em impact map} (i.e., the post-impact velocity prediction) is obtained starting from the standard equations of motion
\begin{align}
    \vM(\vq)\ddot \vq + \vh(\vq, \dot \vq) & = \vtau + \vJ^T_\vN(\vq) \lambda_N,
    & \text{if } g_N(\vq) & = 0, \\
    \vM(\vq)\ddot \vq + \vh(\vq, \dot \vq) & = \vtau ,
    & \text{if } g_N(\vq) & > 0,
\end{align}
where $\vq \in \mathbb{R}^n$ are the generalized coordinates,
$M$ the mass matrix, $\vh$ the Coriolis, centrifugal, and gravity terms, $\vtau$ the actuation torque,
$g_N$ the
{\em gap function} representing the
distance between the robot and the contact surface (obtainable via forward kinematics), $\lambda_N \in \mathbb{R}$ is the normal contact force, and $\vJ_N(\vq) \in \mathbb{R}^{1 \times n}$ is the corresponding contact Jacobian ($\vJ_N(\vq) := \partial g_N/\partial \vq$).
In nonsmooth mechanics, $\lambda_N$ is allowed to become impulsive at the moment of collision and this leads to the so-called {\em impact equation}
\cite{Br16B_nonsmooth}
\begin{align}
    \label{eq:impact_equation}
    \vM(\vq) (\dot \vq^+ - \dot \vq^-)  = \vJ^T_N(\vq) \Lambda_N,
\end{align}
where $\Lambda_N$ represents the impulsive force magnitude
and $\dot \vq^+$ and $\dot \vq^-$ denote the post- and
ante-impact joint velocities, respectively.
The impact map is obtained combining the impact equation with
an {\em impact law} that in case of a frictionless\footnote{\added{For a review
of the state of the art for \emph{partially elastic}
impact maps, including different notions of coefficient of restitution (CoR), the reader
is referred to the recent
\cite[Section 1.2.1]{JiGaMu19J_Batting} and
references therein.}}
inelastic impact as we consider in this work reads
\begin{align}
   \label{eq:impact_law}
   \dot g_N^+ = \vJ_N(\vq) \dot \vq^+ & = 0 .
\end{align}
The equation above is simply stating that the normal component of the
Cartesian post-impact velocity of the end effector
should be zero after the impact.
The impact map allows for a velocity jump, while keeping the configuration unaltered.
The combination of \eqref{eq:impact_equation} and \eqref{eq:impact_law}
allows\footnote{
\added{Multiply \eqref{eq:impact_equation}
on the left by $\vM^{-1}(\vq)$ and then
the result again on the left by $\vJ_N(\vq)$,
taking into account \eqref{eq:impact_law},
finally obtaining
$\Lambda_N = - (\vJ_N(\vq) \vM(\vq)^{-1} \vJ^T_N(\vq))^{-1} \vJ_N(\vq) \dot \vq^-.$
}
}
to express $\Lambda_N$ as a function $\dot \vq^-$
\deleted{obtaining an explicit expression for
$\vq^+$ from \eqref{eq:impact_equation}}
and this \added{fact},
by substituting \replaced{the newly obtained}{this} expression in
\eqref{eq:impact_equation},
to predict the value of $\dot \vq^+$.
One obtains therefore
the following single-point frictionless inelastic {\em impact map}, valid for $g_N(\vq) = 0$,
\begin{align}
  \label{eq:impact_map}
   \dot \vq^+ & =
       \left( \vI - \vM^{-1} \vJ_N^T (\vJ_N \vM^{-1} \vJ_N^T)^{-1} \vJ_N \right) \dot \vq^- .
\end{align}
In the presence of friction, partially elastic and/or multiple
simultaneous impacts, the formulation of the impact map becomes necessarily more sophisticated and often implicit \cite{Gl06J_IntroImpacts} but this is not essential for transmitting the core message of this work and is
therefore left out in this brief overview.


\section{PROBLEM STATEMENT}
\label{sec:probstatement}
\begin{figure}[ht!]
    \centering
    \includegraphics[width=0.475\textwidth]{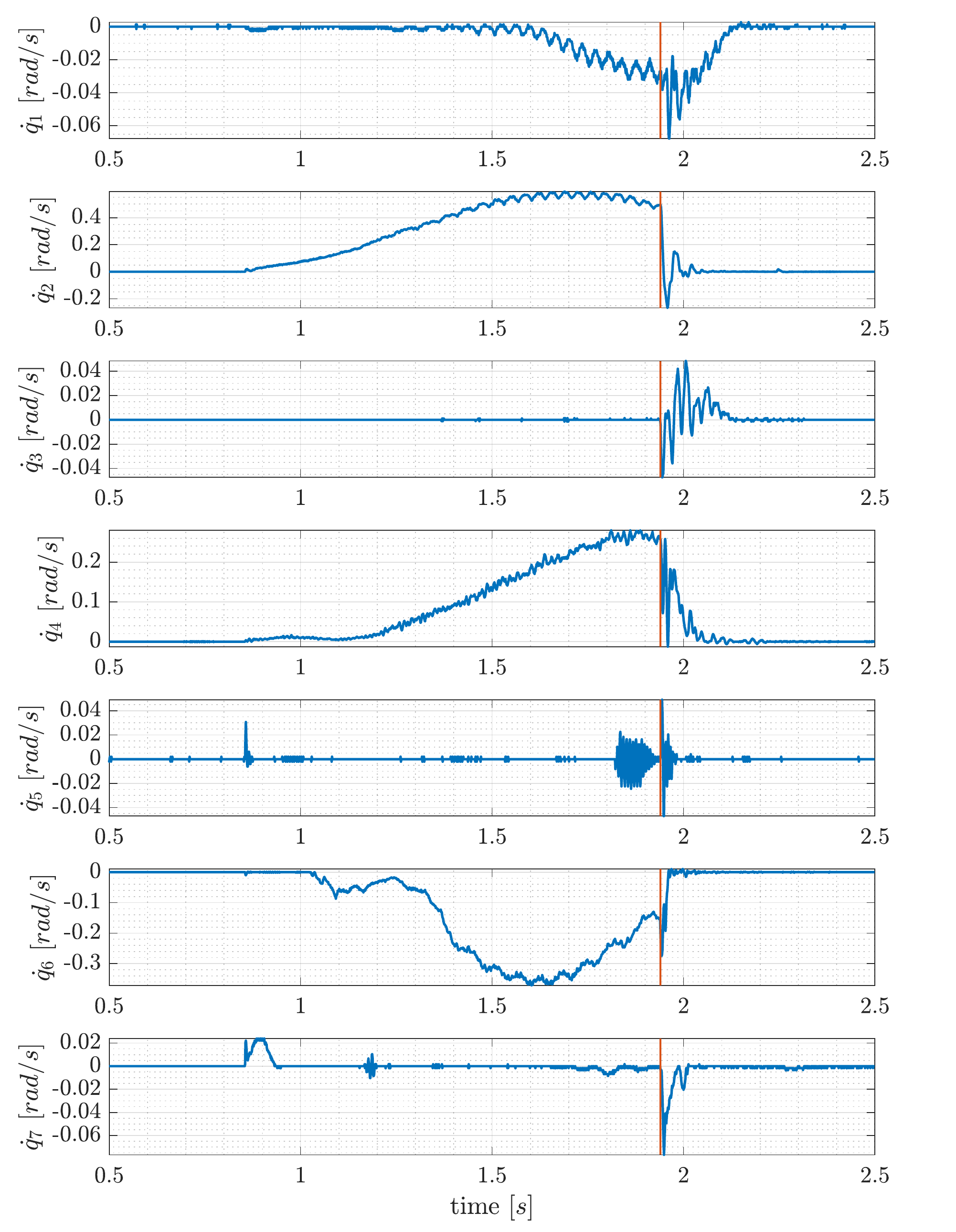}
    \vspace*{-4mm}
    \caption{Joint velocity signals
    for an impact at 0.2~m/s and 90~deg with respect to the horizontal table
    (\added{note different scaling
    for y-axes; }indices from 1 to 3 are for the shoulder, 4 for the elbow, and
    5 to 7 for the wrist\added{ as in Figure~\ref{fig:snap_kuka_exp}}). The vertical red line corresponds to the estimated impact time. Joint velocities are obtained via finite difference of the recorder encoder data.}
    \label{fig:impact_table_0p2_00}
\end{figure}

Several impact experiments have been conducted between a torque controlled robotic arm (KUKA LWR IV$+$) and a smooth \replaced{sturdy}{wooden} table (cf. Figure~\ref{fig:snap_kuka_exp}). The robot impacted the table\added{, made of hard wood,} via a spherical metal probe that was secured with bolts to the standard robot tool mounting plate. A representative example of the recorded joint angles during an impact experiment is given in Figure~\ref{fig:impact_table_0p2_00}. In this particular experiment, the impact between the table and the robot is normal to the table and occurs at a Cartesian velocity of 0.2 m/s, approximately at time 1.94 s. A post-impact damped vibratory response can be observed, lasting approximately 100 ms. Just the second, fourth, and sixth joint are notably affected by the impact: this is justified by the fact that the impact motion occurs essentially on a vertical plane (2D motion), which is also an approximate plane of symmetry for the robot and in which
the mentioned joints are the ones that affect the arm motion the most (essentially, we are looking at a planar impact of a planar RRR manipulator).

Besides this particular experiment, various combinations of low impact velocities and angles have been recorded (about twenty experiments, with repetitions). Overall, the impact velocity varied between 0.1 and 0.2 m/s and the impact angle between 30 and 90 degrees with respect to the table surface.

The impact experiments were obtained employing a task-based QP robot controller\footnote{\url{https://github.com/kuka-isir/rtt_lwr/releases} \newline and \url{https://orca-controller.readthedocs.io/}}
(cf. \cite{LJ18C_Toward_XRAY} and \cite{LJ18C_Experimental_Validation}
and references therein). The control torques are computed based on 7DOF rigid-body kinematics, velocity kinematics and dynamic models obtained using the KDL library\footnote{\url{https://github.com/orocos/orocos_kinematics_dynamics}}\added{, and fed directly at 1 KHz into the low-level joint torque control loop provided by the manufacturer}. The controller is assigned a pose task for the end-effector (with a linear motion for the metal probe center with constant velocity and constant orientation) and a regularization task (constant joint posture) to avoid self motions. The goal pose is located below the wooden table and cannot be reached as impact with the table occurs first. Once contact has been detected using torque measurements at the joint level, the controller switches to pure gravity compensation.

\begin{figure}[ht!]
    \centering
    \includegraphics[scale=0.65]{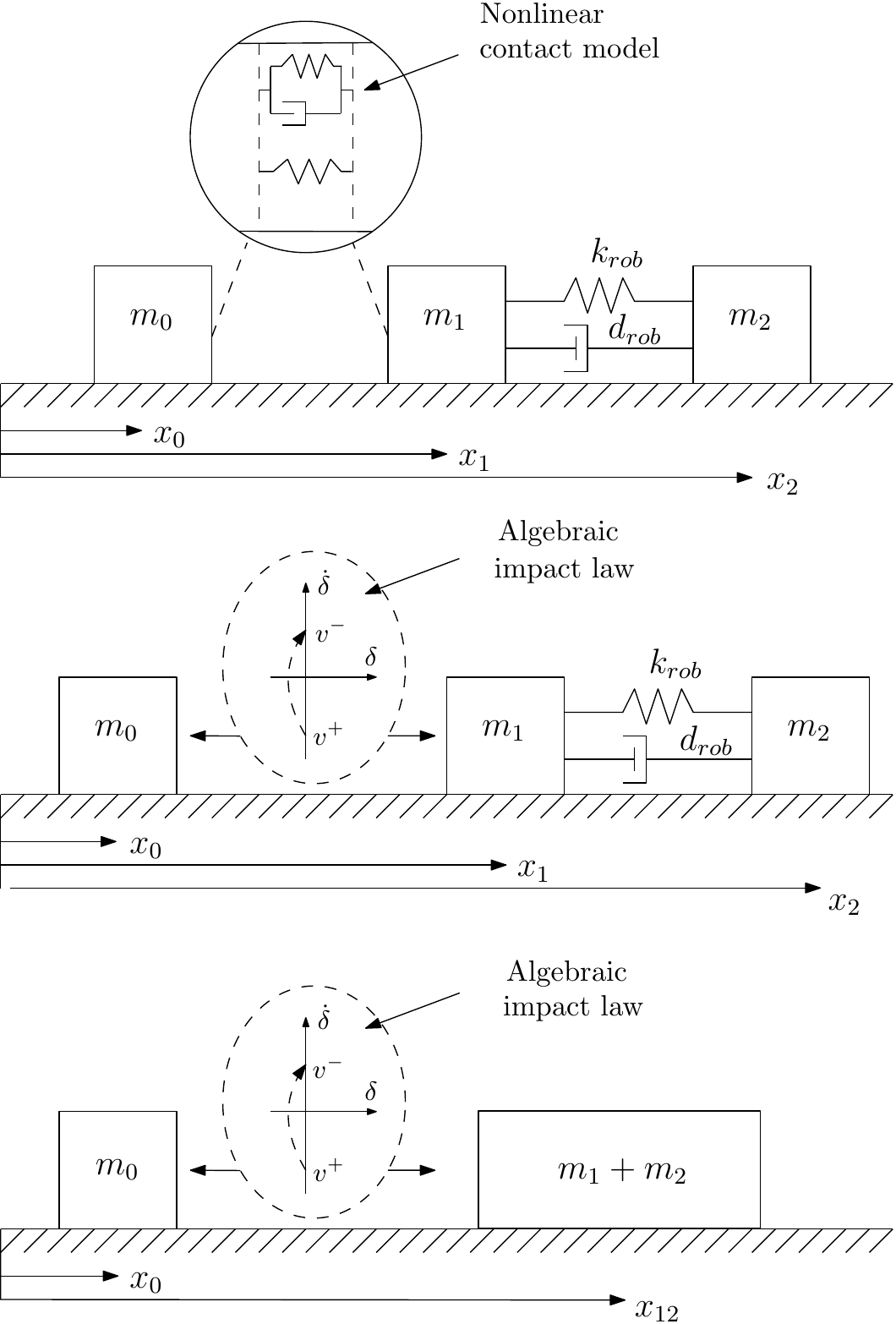}
    \caption{Three levels of abstraction of an impact scenario, where a ``1-DOF robot'' (right) physically interacts with an external rigid object (left): (top) Contact is compliant and the robot is flexible;  (middle) Contact is modelled via an algebraic impact law and the robot is flexible; (bottom) Contact is modelled via an algebraic impact law and the robot is rigid.}
    \label{fig:three_models}
\end{figure}

\begin{figure*}[ht!]
    \centering
    \includegraphics[scale=0.58]{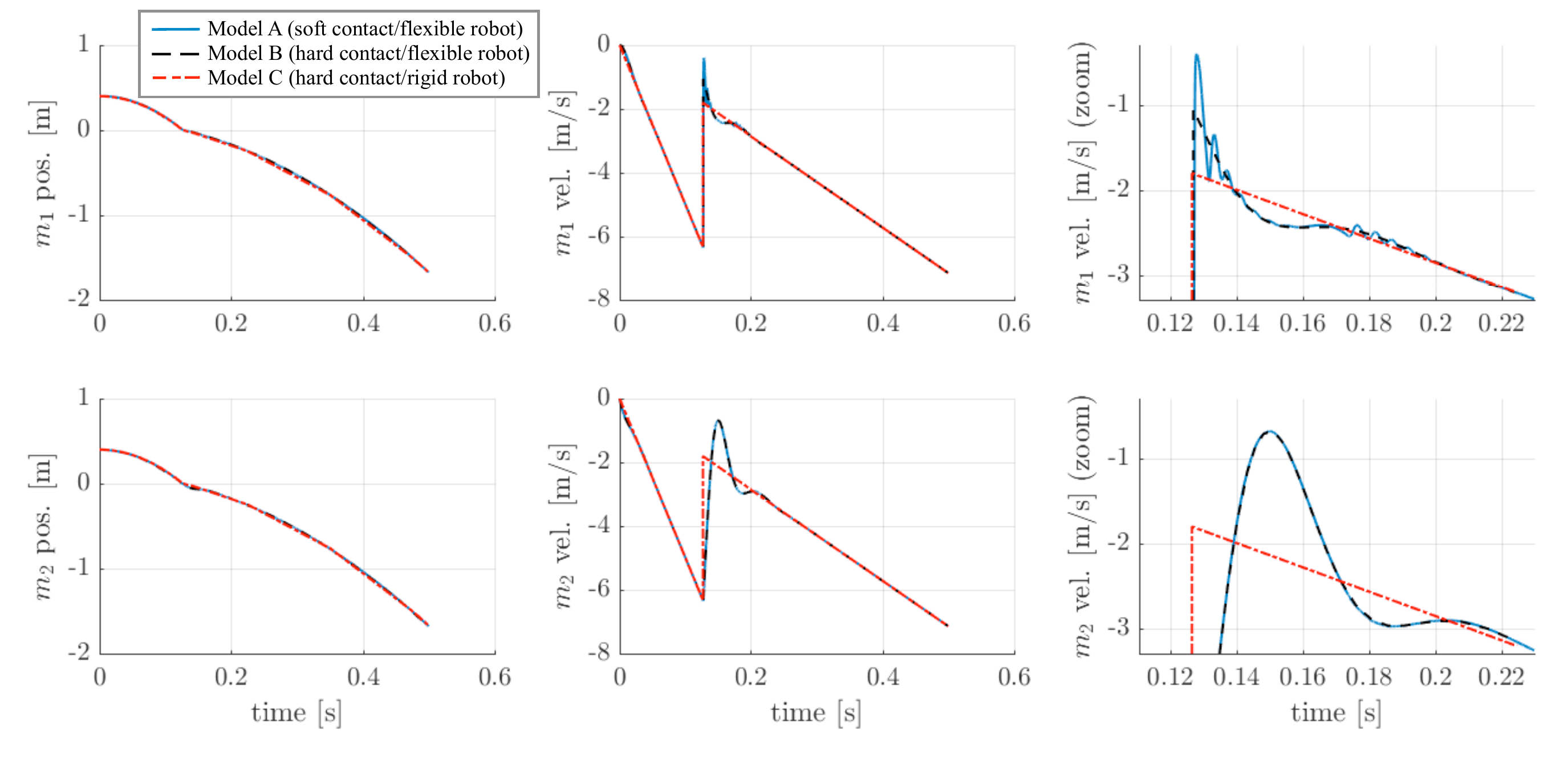}
    \caption{Simulation results corresponding to the three models,
    where motion signals are given for mass $m_1$ and $m_2$.
    The first column shows the position of mass $m_1$ (top) and $m_2$ (bottom).
    The second column shows the velocity of mass $m_1$ (top) and $m_2$ (bottom).
    The third column is
    a zoomed version of the second column, about the impact time.
    Model A (compliant-contact/flexible-robot) is depicted in blue, Model B
    (hard-contact/flexible-robot) in dashed black, and
    Model C (hard-contact/rigid-robot)
    in dash-dot red. }
    \label{fig:three_model_response}
\end{figure*}

To validate the lack of influence of the \added{user-defined} controller in the impact response, different experiments have been performed where the Cartesian task PD gains \added{of the user-defined controller} have been altered \replaced{confirming}{verifying} that no appreciable difference in the post-impact response in the joint signals could be observed. This leads to hypothesize that the post-impact oscillations (cf. Figure~\ref{fig:impact_table_0p2_00}) are solely
of structural nature or due to the high bandwidth low-level torque controllers active on the joints\added{, which is developed
by the robot manufacturer and encoded in the robot firmware}.
A confirmation of this \replaced{hypothesis}{fact} would require the mounting of accelerometers and accessing both motor and joint encoders on the robot: this is deemed as a future research but the observed oscillations are very likely due to joint level non-rigidities induced by the gears (Harmonic drive) of the transmission as well as the joint torque control technology.  
What is relevant for the discussion that follows is, in any case, that the system exhibits damped oscillatory modes whose time scales are of at least an order of magnitude higher than the impact phenomenon.

As anticipated in the introduction, the impact data reported in Figure~\ref{fig:impact_table_0p2_00} makes it apparent that there is a fundamental challenge when trying to employ a post-impact velocity prediction based on a rigid-body robot model. Post-impact predictions based on rigid-body models do not exhibit any oscillatory behavior after an impact and therefore it is unclear how their prediction can be validated against real impact experiments that present damped post-impact oscillatory transients.
The challenge is therefore summarized in the following problem statement, for which we propose a solution
in Section~\ref{sec:contribution}.


\noindent\textbf{Problem statement.} \emph{How can experimental post-impact velocity data be quantitatively compared with post-impact velocity predictions, readily obtainable via an available rigid-body robot model and an algebraic impact law?}

Note that post-impact velocity predictions based on rigid-body models
are essentially what dynamic simulators (e.g., Bullet, ODE, or Vortex Studio) embedded in mainstream robot simulation suites such as Gazebo or CoppeliaSim (former V-REP) provide, at least in simple impact scenarios.


\section{CONTRIBUTION}
\label{sec:contribution}

In this section, we propose \added{a
thinking framework and} a procedure to assess the performance
of the post-impact prediction obtainable with a rigid body impact model against the recorded
impact data.

\noindent
\textbf{\added{Thinking framework.}} At first, for illustration purposes, this \added{thinking framework and} procedure
is introduced by means of an academic example
employing two bodies colliding with each other along a straight line.
We model these two bodies and the interaction between them in different ways,
obtaining in total three different dynamical systems,
as depicted in Figure~\ref{fig:three_models}.
The purpose of this
exercise
is to
\added{introduce the new thinking framework and}
show that a fully rigid model with nonsmooth impact law
is capable of capturing the ``steady state'' response of more accurate and realistic flexible models and that
by removing the flexibility-induced transient response from the flexible models, the rigid model and the flexible ones can be quantitatively compared at impact time.

All these three models
are made of an ``interaction environment'' (the body depicted on the left,
considered fully rigid and with mass $m_0$)
and a ``robot'' (the body depicted on the right, with total mass $m_1 + m_2$). The interpenetration
and relative velocity
between the interaction environment and the body
are denoted, respectively, as $\delta$
and $v$. Note that $\delta = -g_N$, with $g_N$
the gap function as in Section~\ref{sec:nonsmooth}.
The three models are detailed in the following bullets.

\noindent\textbf{$\bullet$ Model A: Compliant-contact/flexible-robot.} In this model,
the robot (right) is assumed to be flexible and represented as the ensemble of two masses ($m_1$ and $m_2$) connected via a compliant coupling, with linear stiffness $k_{rob}$ and damping $d_{rob}$. The interaction between the robot and the environment (the mass $m_0$ on the left) is modeled via the Hunt-Crossley nonlinear contact model
\cite{HuCr75J_CoRasDamping}, which relates the interpenetration $\delta$ and its rate of
change to the contact force $F$. Namely, we have
\begin{align}
\label{eq:HCforce}
F(\delta, \dot \delta) =
 \begin{cases}
      k_{env}~\delta^c + d_{env}~\delta^c ~ \dot{\delta},     &  \delta \geq 0, \\
      0,                                                    &  \delta < 0,
   \end{cases}
\end{align}
with $k_{env}$, $d_{env}$, and $c$ chosen constants.
Due to the interpenetration-dependent damping coefficient $d_{env}~\delta^c$,
the Hunt-Crossly contact model does not exhibit any (nonphysical) contact force jump when contact is established at non-zero velocity. In the simulations presented hereafter in this section,
the damping coefficient $d_{env}$ is set to a very high value in order to simulate (almost) inelastic impacts;

\noindent\textbf{$\bullet$ Model B: Hard-contact/flexible-robot.} In this model,
the contact interaction between the (flexible) robot is modelled via the fully inelastic impact law
$\dot\delta^+ = {\dot x}_1^+ - {\dot x}_0^+ = 0$, triggered whenever $\delta := x_1 - x_0 = 0$
and $\dot\delta^- = {\dot x}_1^- - {\dot x}_0^- < 0$ (``$+$'' and ``$-$'' denote left and right limits at impact time);

\noindent\textbf{$\bullet$ Model C: hard-contact/rigid-robot.} In this model,
the robot is considered as a point mass (with mass  $m_1 + m_2$) and
the robot-environment contact is rigid (same impact law as in Model B).

Looking at these three models, it should be apparent that model A is the closest one to physical reality, while model C, based on rigid-body assumption, represents the one that is typically available and used for robot control and planning.

%
~\\
\noindent\textbf{Illustrative numerical simulation of an impact.}
We consider the situation where the environment ($m_0$) and the robot ($m_1$ and $m_2$) start at rest and a constant force is applied to $m_2$. This force accelerates the robot towards the left until it impacts with the environment and a sudden velocity change is experienced, making the environment start moving and the robot to suddenly decelerate.
In Figure~\ref{fig:three_model_response}, we report the corresponding motion of the robot
for the three models described above (and parametrized using the values provided in Table~\ref{tab:parasim}), by showing the position and velocity signals of the two masses $m_1$ and $m_2$ (for model C, the motion of $m_1$ and $m_2$ are identical as they are rigidly connected).
\begin{table}[ht!]
\centering
\caption{Parameters used in the numerical simulations}
    \begin{tabular}{l|l|l||l|l|l}
    Parameter & Value & Units & Parameter & Value & Units    \\ \hline
    $m_0$     & 5     & kg     & $m_1$    & 1     & kg       \\
    $m_2$     & 1     & kg     & $c$      & 3/2    & [-]     \\
    $k_{env}$ & $1\cdot10^8$   & $N/m^c$    & $d_{env}$ & $1\cdot10^8$     & $Ns/m^{c+1}$ \\
    $k_{rob}$ & $1\cdot10^4$   & N/m      & $d_{rob}$ & 80     & N s/m
    \end{tabular}
    \label{tab:parasim}
\end{table}
Without loss of generality, the initial separation between $m_0$ and $m_1$ was set to $0.4$ m and the constant pushing force to $100$ N. The simulation parameters for the Hunt-Crossley model ($k_{env}$, $d_{env}$, and $c$) have been chosen
based on the metal-wood interaction in the real robot-table experiment,
assuming a stiffness corresponding to hard wood, a sphere-halfspace Hertz contact,
and high damping to represent an impact with a small coefficient of restitution. The robot stiffness   $k_{rob}$ is chosen to be lower than the contact's and damping $d_{rob}$ is selected to get a lightly damped response\added{ (similar
results can be obtained for different value of $d_{rob}$ as long as the settling time does not become unreasonably high, as it can be readily verified via the provided MATLAB scripts)}. Simulations were performed in MATLAB using
standard ODE solvers (a variable-order stiff solver,
\texttt{ode15s}, is used to deal with the stiff ODE related to high contact stiffness used in model A).

\noindent\textbf{Discussion about the simulation results --}
The numerical results on Figure~\ref{fig:three_model_response} show that the dynamic responses of robot in models A and B are essentially indistinguishable at the time scale of interest.
This is simply an illustration of the time-and-space scale separation of contact and body dynamics, which justify the contact modeling simplification employed in nonsmooth mechanics.
The zoomed-in version of the impact response of $m_1$ depicted in the last column of Figure~\ref{fig:three_model_response} shows that the compliant and nonsmooth rigid contact models do indeed differ on a millisecond time scale, with the nonsmooth model B just capturing the average response of the compliant model A. The
numerical simulation of model A shows explicitly two distinctive dynamics at work: one fast due to contact stiffness ($0.12-0.14$ s in the top subplot) and the one slower, captured also by model B ($0.12-0.21$ s and beyond).
Inspection of the gap function actually shows bouncing of mass $m_1$ on $m_0$ before full adhesion, a phenomenon that is
comparable with foot chattering in legged locomotion \cite{Al90J_ThreeUsesSpring}.
Other simulations with higher damping at contact (namely, $1 \cdot 10^9$ $Ns/m^{c+1}$), instead, show no bouncing between
$m_1$ and $m_0$ and an even closer matching
between the motions of mass $m_1$ for models A and B
(these plots\added{, including those that can be generated by perturbing the damping $d_{rob}$,} are not reported here for space limitation, but they are easily reproduced with the provided MATLAB script).

Interesting to note is that the responses of model A and model B at the ``actuation-and-position-sensing'' side (i.e., $m_2$) are essentially indistinguishable, as clearly shown in the bottom middle and right plots of Figure~\ref{fig:three_model_response}.

\begin{figure}[ht!]
    \centering
    \includegraphics[scale=0.52]{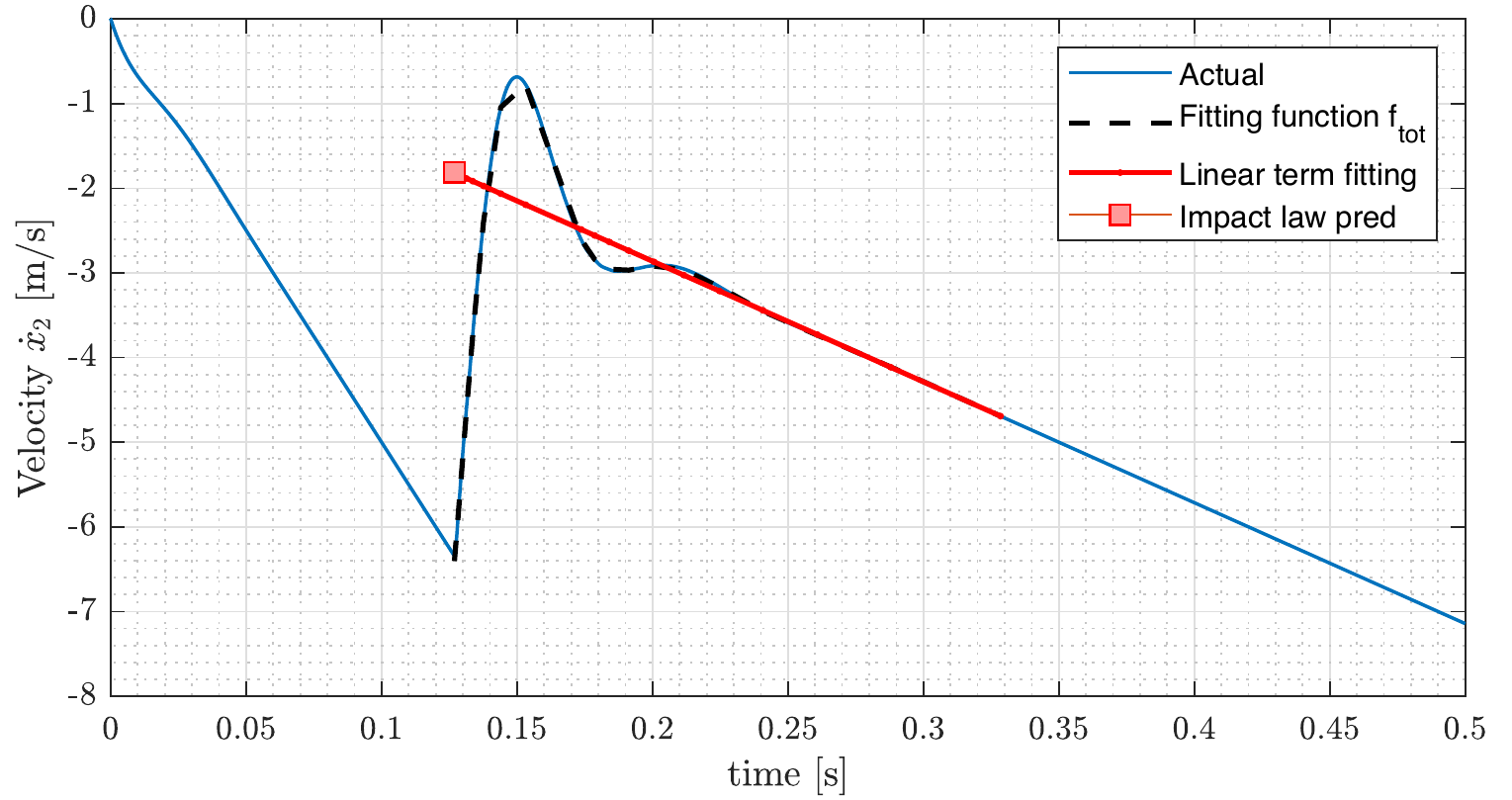}
    \caption{Numerical data fitting result. Here we used the fitting scheme on the velocity signal of mass $m_2$, where the linear term of the fitting function is extrapolated back to the time instance of impact. In addition, the prediction of the post-impact velocity of the impact map is also provided, which is compared to the estimation provided by the extrapolation.}
    \label{fig:num_fitting}
\end{figure}

\noindent\textbf{Key observation \deleted{and contribution}.} At this point, it becomes interesting also to
compare the behavior of model B (flexible robot) with that of model C (rigid robot). In particular, regarding the actuation-and-sensing side of the robot ($m_2$).
Figure~\ref{fig:three_model_response} illustrates that the velocity of the flexible robot model converges to that of the rigid robot model in about 100 ms, suggesting to interpret the response of the flexible robot model as that of a low-pass filter acting on a velocity step input.
The flexible robot dynamics after an impact are thus seen as a faster dynamics than that of the gross motion of the bodies after impact. Given the time-and-space scales of these vibrations (in real experiments, these vibrations are only observable looking at encoder data), we suggest to treat the post-impact vibration dynamics as linear and
thus as the superposition of exponentially decaying oscillatory responses (including constant and linear terms corresponding to a step input). Least-squares fitting/modal analysis techniques can thus be
used to remove the oscillatory part of the post-impact time response, extracting just the steady-state response (sum of a constant and a linear term as a function of time). 
An example of application of this procedure, where the velocity signal is decomposed into the sum of a constant, a linear function of time, and just one exponentially decaying function of the form $A \exp( \omega t + \phi )$ is given in Figure~\ref{fig:num_fitting}.

\noindent\textbf{Proposed quantitative comparison procedure.}
Summarizing, we propose the following method to
quantitatively compare experimental post-impact data
with post-impact velocity predictions
derived from a rigid multi-body impact map:
\begin{enumerate}
  \item Identify the impact time $t_i$ in the experimental data by looking at, e.g., sharp variations
    of joint encoder data (either visually as we did in this work or by the use of automatic methods such as
    those described in \cite{RiSaNi18J_JAfilter} and references therein);
  \item Extract the impact robot configuration $\vq(t_i)$ and   corresponding pre-impact
    joint velocity ${\dot \vq}^-(t_i)$ and
    compute the rigid-robot post-impact joint velocity estimate $\overset{\Delta}{\dot \vq}\hspace{0mm}^+(t_i)$
    employing the impact map
    \eqref{eq:impact_map};
  \item Use (nonlinear) least squares fitting
  or frequency-domain-based procedures on the signal $\dot \vq(t)$ to
  separate the affine (=constant plus linear) response from the oscillatory damped response (``sum of
  eigenmodes'') over the interval
  $[t_i, t_i + T_s]$
  where vibrations are observed (with $T_s$ the settling time).
  Employ the affine part to
  construct the \emph{virtual rigid-robot post-impact velocity} $\widehat{\dot \vq}^+(t_i)$;
  \item Evaluate the (relative and absolute) error between $\overset{\Delta}{\dot \vq}\hspace{0mm}^+(t_i)$
  and $\widehat{\dot \vq}^+(t_i)$ for impacts occurring at different postures and velocities,
  to quantify the general accuracy of the post-impact velocity estimation (considering the virtual rigid-robot post-impact velocity $\widehat{\dot \vq}^+(t_i)$ the ground truth obtained from experiments).
\end{enumerate}
The procedure above can be employed in Cartesian space, rather than in joint space, by performing points 3 and 4
using the Cartesian velocities
$\overset{\Delta}{\vv}\hspace{0mm}^+(t_i)
 = \vJ(\vq(t_i)) \overset{\Delta}{\dot \vq}\hspace{0mm}^+(t_i)$
and
$\widehat{\vv}^+(t_i)$, obtained by least squares fitting
on the signal $\vJ(\vq(t)) {\dot \vq}(t)$,
in place of the joint velocities
$\overset{\Delta}{\dot \vq}\hspace{0mm}^+(t_i)$ and $\widehat{\dot \vq}^+(t_i)$, respectively.
The result of applying this method on the 7DOF KUKA arm
are reported in the following section.



\section{FITTING RESULTS ON THE 7DOF ARM}
\label{sec:results}




\begin{figure*}[!ht]
     \centering
    \includegraphics[width=0.88\textwidth]{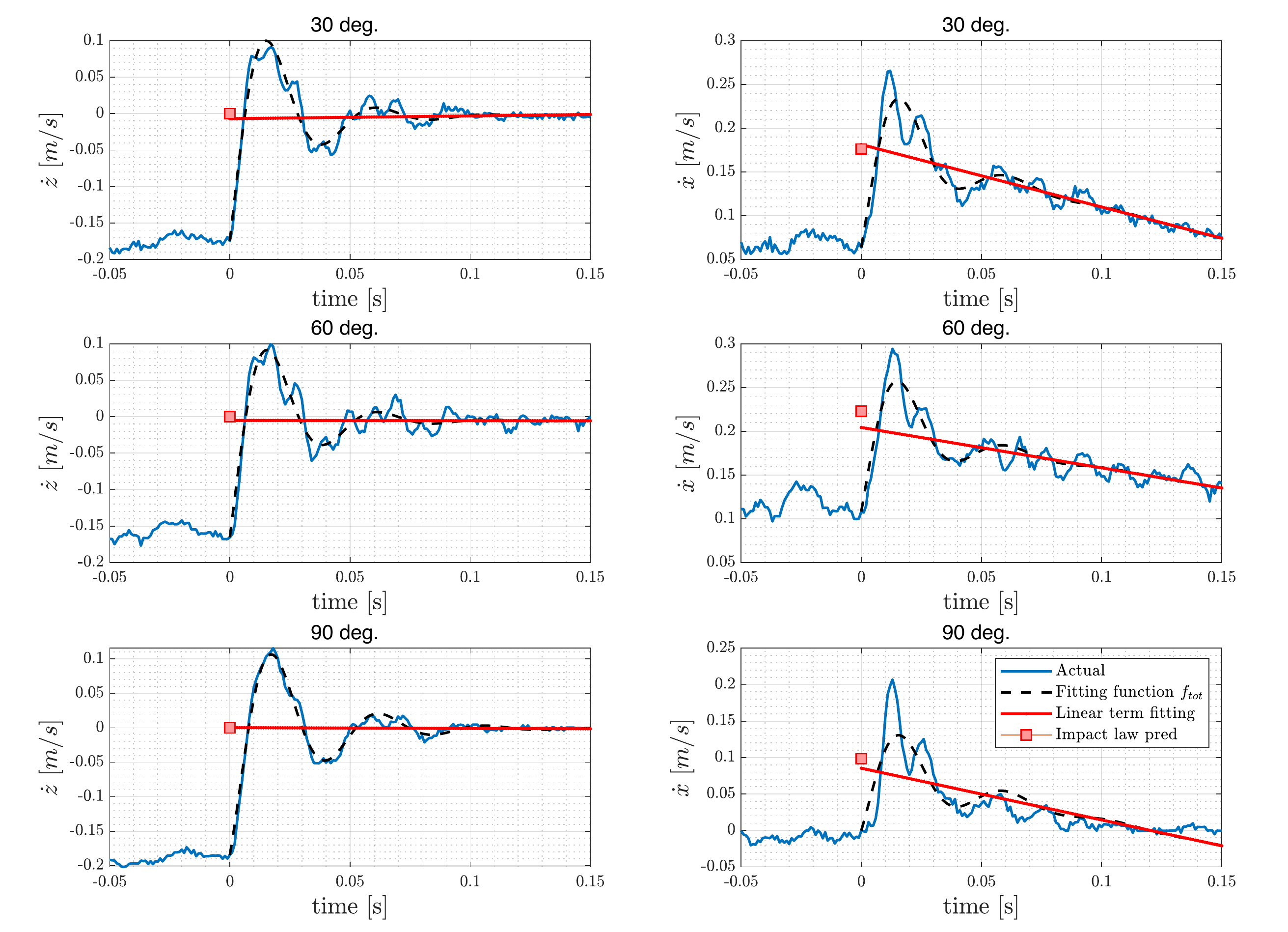}
    \vspace*{-4mm}
    \caption{Cartesian velocity fitting results for impact at 0.2 m/s under angles ranging from 30 to 90deg w.r.t. to the horizontal table. The x-direction (right) is tangential to the table and the z-direction (left) is normal.
    Note how the fully rigid-model impact law prediction (red square) agrees with the prediction resulting from filtering out the post-impact transient (left end point of the red segment)}
    \label{fig:resfit0p2}
\end{figure*}

Figure~\ref{fig:resfit0p2} illustrates the result of the fitting procedure
described in the previous section
now applied to the 7DOF arm experimental data
collected for an impact at 0.2 m/s and three different impact angles.
Only vertical and longitudinal Cartesian velocities are reported, as lateral displacement is negligible in the performed impact experiments as explained in Section~\ref{sec:probstatement}. Cartesian velocities are obtained from the recorded joint velocities and  end-effector Jacobian.
For least squares fitting, we employ for each Cartesian velocity component the fitting function
\begin{align}
  f_{tot}(t; \vp)
  & :=
  v^- + a t + A  \left( e^{\gamma t} \cos (\omega t + \phi) - \cos(\phi) \right)
\end{align}
where $t$ denotes the time after the impact $t_i$,
$v^- \in \mathbb{R}$ the pre-impact velocity,
and $\vp := (a, A, \gamma, \omega, \phi) \in \mathbb{R}^5$
denotes the vector of parameters (slope $a$,  amplitude $A$, decay factor $\gamma$, frequency $\omega$, and phase shift $\phi$)
for the
least-squares fitting.

More precisely, the full set of parameters is only used
for the vertical Cartesian direction (which has always a clear dominant second order response with large amplitude)
while for the fitting of the motion in the longitudinal Cartesian direction the frequency and decay rate are set  equal to the one identified for the vertical Cartesian direction. In this way, we obtain a single real eigenmode to describe the oscillation in accordance with the post-impact linear oscillation assumption.
The fitting procedure is applied on a 150 ms time window, that was selected
based on the stabilization of the least squares fitting parameters
and roughly corresponds to three oscillation periods.
In Figure~\ref{fig:resfit0p2}, the reconstructed
virtual rigid-body response $v^- - A \cos(\phi) + a t$
is shown as a red line (left $z$, right $x$ Cartesian component). Its value
${\hat v}^+ = v^- - A \cos(\phi)$ at impact time
(the tip of the red line)
should be compared
with the post-impact velocity estimate $\overset{\Delta}{v}\hspace{0mm}^+$
(the red square),
derived via the rigid-body impact map. The impact map is derived from the rigid-body robot model employed by the QP-based robot controller, combined with
the frictionless inelastic impact law between the end-effector tip
and the (assumed rigid) wooden table, as discussed in Section~\ref{sec:nonsmooth}\added{, specifically in
\eqref{eq:impact_equation} and \eqref{eq:impact_law}}.

{\setlength{\tabcolsep}{.5em}
\begin{table*}
\centering
   \caption{Evaluation of the impact map and fitting results for the two experimental data sets (data in $m/s$)}
   \label{tab:eval1}
    \begin{tabular}{l|llllllll|llllllll}
    velocity, angle
    &
    $ \tilde{v}^-_z$  & $\overset{\Delta}{v}\hspace{0mm} ^+_z$ & $\hat{v}^+_z$ & $\eta_{z}$
    & $\tilde{v}^-_x$ & $\overset{\Delta}{v}\hspace{0mm} ^+_x$ & $\hat{v}^+_x$ &
    $\eta_{x}$
    &
    $ \tilde{v}^-_z$  & $\overset{\Delta}{v}\hspace{0mm} ^+_z$ & $\hat{v}^+_z$ & $\eta_{z}$
    & $\tilde{v}^-_x$ & $\overset{\Delta}{v}\hspace{0mm} ^+_x$ & $\hat{v}^+_x$ &
    $\eta_{x}$
    \\
    \hline\hline %
  0.10 m/s, 30$^\circ$  & -0.097 & 0 & -0.006 & 0.006 & \phm0.031 & 0.094 & 0.096 & 0.002  & -0.085 & 0 & -0.007 & 0.007 & 0.061 & 0.120 & 0.112 & 0.008 \\
  0.10 m/s, 60$^\circ$  & -0.085 & 0 & -0.009 & 0.009 & \phm0.061 & 0.122 & 0.109 & 0.013  & -0.099 & 0 & -0.003 & 0.003 & 0.028 & 0.088 & 0.094 & 0.006 \\
  0.10 m/s, 90$^\circ$  & -0.101 & 0 & -0.005 & 0.005 & \phm0.004 & 0.055 & 0.050 & 0.006  & -0.103 & 0 & -0.003 & 0.003 & -0.007 & 0.039 & 0.050 & 0.011 \\
  0.15 m/s, 30$^\circ$  & -0.175 & 0 & -0.007 & 0.007 & \phm0.064 & 0.176 & 0.181 & 0.005  & -0.120 & 0 & -0.012 & 0.012 & 0.092 & 0.178 & 0.170 & 0.008 \\
  0.15 m/s, 60$^\circ$  & -0.114 & 0 & -0.010 & 0.010 & \phm0.096 & 0.174 & 0.163 & 0.011  & -0.137 & 0 & -0.006 & 0.006 & 0.054 & 0.140 & 0.149 & 0.009 \\
  0.15 m/s, 90$^\circ$  & -0.148 & 0 & -0.003 & 0.003 & -0.006 & 0.072 & 0.064 & 0.008 & -0.148 & 0 & -0.002 & 0.002 & 0.002 & 0.072 & 0.072 & 0.000 \\
  0.20 m/s, 30$^\circ$  & -0.175 & 0 & -0.007 & 0.007 & \phm0.064 & 0.176 & 0.181 & 0.005  & -0.144 & 0 & -0.011 & 0.011 & 0.120 & 0.222 & 0.209 & 0.013 \\
  0.20 m/s, 60$^\circ$  & -0.165 & 0 & -0.005 & 0.005 & \phm0.108 & 0.223 & 0.204 & 0.019  & -0.168 & 0 & -0.004 & 0.004 & 0.086 & 0.190 & 0.190 & 0.000 \\
  0.20 m/s, 90$^\circ$  & -0.185 & 0 & \phm0.000 & 0.000 & \phm0.000 & 0.098 & 0.085 & 0.013   & -0.178 & 0 & -0.002 & 0.002 & 0.003 & 0.093 & 0.099 & 0.006
  \end{tabular}
\end{table*}
}

Overall, the reconstructed and rigid-body impact map predictions, ${\hat v}^+_i$ and
$\overset{\Delta}{v}\hspace{0mm}^+_i$, $i = \{x, z\}$,
are in very good agreement.
As summarized in Table~\ref{tab:eval1}, this holds not
just for the $0.2$ m/s
impacts, but also for the impact experiments at $0.15$ m/s and $0.1$ m/s.
The table shows both the normal and longitudinal velocity
as well as the absolute prediction error
$\eta_i := | \overset{\Delta}{v}\hspace{0mm}^+_i - \hat{v}^+_i |$,  $i = \{x, z\}$, in $m/s$.
On average, we get a 8 mm/s absolute error and
7.3\% relative error on predicting post-impact sliding velocity
(the relative error is computed as
$2 | \overset{\Delta}{v}\hspace{0mm}^+_i - \hat{v}^+_i |
/ | \overset{\Delta}{v}\hspace{0mm}^+_i + \hat{v}^+_i |$),
$i = \{x, z\}$.

Looking at the measured impact response, it is noticeable that
the assumption of a second order system type response is only partially valid and higher frequency modes currently not modeled are present. This is more noticeable at the joint level (results are reproducible with the provided MATLAB scripts) and therefore we have chosen here to limit the model analysis based on a single mode of vibration at Cartesian level. The use of a more sophisticated modal analysis procedure at joint level is considered as the next step of our investigation.

\section{CONCLUSIONS}
\label{sec:conclusions}


Several impact experiments between a 7DOF torque-controlled robotic arm and a \replaced{sturdy}{wooden} table
have been performed, at different velocities and impact angles. The proposed post-impact velocity prediction procedure,
based on the idea of removing post-impact oscillation components,
shows to be in very good agreement with
a fully-rigid robot impact model.
This procedure, which is not tailored to a specific robot, can be used to assess if a fully rigid-body impact model can provide reliable post-impact velocity prediction that are of use in impact-aware robot planning, control, and perception.

For the specific impact experiments performed with the KUKA
LWR IV+, we obtained a 7.3\% relative error and 8 mm/s absolute error difference between the measured and predicted sliding velocity, over a set of 18 experiments. This good accuracy is rather surprising giving the fact that we are comparing an ideal rigid-body impact model with real experiments that contain unmodeled post-impact (structural) vibrations, that are ``filtered out'' by the proposed procedure.
Further investigation is required to assess
what level of accuracy is required
to achieve satisfying performance in impact-aware manipulation
for specific applications.


For the considered impact scenarios, the post-impact response in the vertical Cartesian direction clearly shows a dominant second-order-system type response to an impact.
On the horizontal direction, the response is more complex and more advanced modal analysis techniques  could be explored
to assess if this could have an effect in the further reduction of the prediction error. We
expect that least-square fitting procedures with multiple oscillatory modes can be borrowed from the structural dynamics literature to this end.

Further research will consider
explicit inclusion of the joint flexibility and motor dynamics in the impact laws, several impact configurations and their influence in post-impact oscillations, different type of robot manipulators (possibly with additional or different compliant components, other than harmonic drives),
oblique impacts with rough surfaces
to study the post-impact velocity predictability in the present of surface friction, and complex impacts such as, e.g., simultaneous grabbing and lifting of heavy boxes with dual-arm systems, with the goal of employing these validated impact models in impact-aware robot manipulation schemes.

\section*{ACKNOWLEDGMENTS}

The authors want to thank the Institut des Syst\`emes Intelligents et de Robotique (ISIR) at Sorbonne Universit\'e and CNRS (Paris, France) for providing access to the KUKA LWR IV+ robot used in this study, Lucas Joseph for collecting the collision data, and Claude Lacoursi\`{e}re, Maarten Jongeneel, and Wouter Weekers for
valuable feedback on a preliminary draft of this document.

\end{document}

%% file: mathdef.tex
\newcommand{\vtau}{\mbox{\boldmath $\tau$}}

\newcommand{\vh}{\mathbf h}

\newcommand{\vp}{\mathbf p}
\newcommand{\vq}{\mathbf q}

\newcommand{\vv}{\mathbf v}

\newcommand{\vI}{\mathbf I}
\newcommand{\vJ}{\mathbf J}

\newcommand{\vM}{\mathbf M}
\newcommand{\vN}{\mathbf N}

